# Saliency-Aware Diffusion Reconstruction for Effective Invisible Watermark Removal

Inzamamul Alam
Department of Computer Science and Engineering
Sungkyunkwan University
Suwon, Republic of Korea
inzi15@g.skku.edu

Md Tanvir Islam
Department of Computer Science and Engineering
Sungkyunkwan University
Suwon, Republic of Korea
tanvirnwu@g.skku.edu

Simon S. Woo*
Department of Computer Science and Engineering
Sungkyunkwan University
Suwon, Republic of Korea
swoo@g.skku.edu

## Abstract

As digital content becomes increasingly ubiquitous, the need for robust watermark removal techniques has grown due to the inadequacy of existing embedding techniques, which lack robustness. This paper introduces a novel Saliency-Aware Diffusion Reconstruction (SADRE) framework for watermark elimination on the web, combining adaptive noise injection, region-specific perturbations, and advanced diffusion-based reconstruction. SADRE disrupts embedded watermarks by injecting targeted noise into latent representations guided by saliency masks although preserving essential image features. A reverse diffusion process ensures high-fidelity image restoration, leveraging adaptive noise levels determined by watermark strength. Our framework is theoretically grounded with stability guarantees and achieves robust watermark removal across diverse scenarios. Empirical evaluations on state-of-the-art (SOTA) watermarking techniques demonstrate SADRE's superiority in balancing watermark disruption and image quality. SADRE sets a new benchmark for watermark elimination, offering a flexible and reliable solution for real-world web content. Code is available on https://github.com/inzamamulDU/SADRE.

## 1 Introduction

Watermarking has long been an essential element in protecting digital assets [5], offering an effective method of ensuring copyright and verifying web content. However, the proliferation of adversarial applications, such as watermark removal [13, 22] has spurred significant interest in developing robust techniques that ensure minimal collateral damage to the underlying content, while effectively disrupting embedded watermarks. This need arises from the inadequacy of existing embedding techniques, which lack robustness and are vulnerable to such adversarial manipulations [4, 22]. Current watermark removal approaches often face a trade-off between the effectiveness of removal and the fidelity of the restored image [4], making it challenging to achieve both goals simultaneously. These approaches typically rely on heuristic-based filtering [15] or hand-crafted features [14]. With the rise of deep learning methods, various data-driven methods treated the watermark removal as an image-to-image translation task [18, 22], although other methods [8, 10] considered both watermark localization and removal in multi-task learning.

With the rise of deep learning techniques across domains [2, 3, 6, 7], researchers have also adopted it for watermark removal [10, 16], which have shown promise in recent years. However, they are often designed for specific watermarking patterns, limiting their generalization to unseen watermarking schemes. Additionally, many of these methods prioritize watermark removal at the expense of image fidelity, resulting in perceptual degradation.

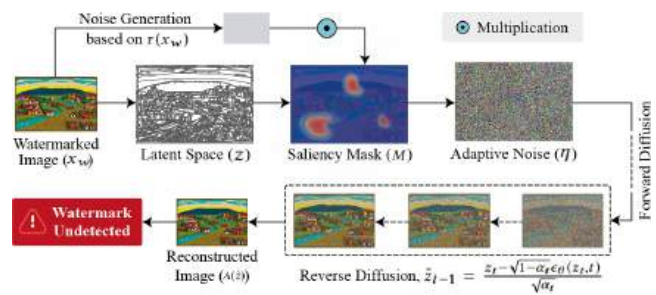

**Figure 1: Overview of the proposed Saliency-Aware Diffusion Reconstruction (SADRE) framework for watermark removal.**

This work proposes a novel salience-aware diffusion reconstruction (SADRE) framework that addresses this challenge by integrating adaptive noise injection with diffusion-based reconstruction. Our method leverages latent space representation to encode the watermark image and injects strategically designed noise to disrupt the watermark while preserving essential image features. To ensure a high-quality reconstruction of the original image, we employ a reverse diffusion process that iteratively removes noise, while

---

maintaining the structural and perceptual fidelity of the image. A key innovation of our approach lies in the adaptive noise injection mechanism, which dynamically adjusts the noise level based on the strength and characteristics of the watermark. By incorporating various noise distributions, such as Laplace, Cauchy, and Poisson, we provide a flexible and robust solution capable of handling various watermarking scenarios. Furthermore, the theoretical underpinnings of our framework, grounded in Hölder continuity [11] and stability guarantees, ensure robust performance under non-linear distortions introduced by watermark embedding.

Empirical evaluations demonstrate the effectiveness of our proposed SADRE across multiple benchmarks, achieving state-of-the-art (SOTA) watermark removal, by preserving image quality.

Thus, the proposed SADRE bridges the gap between theoretical robustness and practical effectiveness, making it a compelling solution for real-world applications.

## 2 Threat Model and Problem Statement

To design an effective attack balancing watermark removal and image fidelity and usability, the following subsection defines the threat model and our problem.

### 2.1 Threat Model

The adversary is assumed to have access to the watermarked image $x_w$, which contains an embedded watermark, potentially used for copyright protection or traceability. The goal of the adversary is to effectively remove or disrupt this watermark such that it becomes undetectable by automated systems or human observers, preserving the visual fidelity of the original image $x$. The adversary operates under the following assumptions:

- The adversary does not possess the clean image $x$, but can estimate the strength of the watermark, denoted by $\tau(x_w)$.
- The adversary does not know the exact watermark embedding mechanism, but assumes that the watermark influences specific regions of the latent representation $z$.
- The adversary aims to balance watermark removal (minimizing detectability), while preserving the image quality.

### 2.2 Problem Statement

Given a watermarked image $x_w$, the objective is to disrupt the watermark such that the reconstructed image $\hat{x}$ is indistinguishable from the original clean image $x$ to both perceptual metrics and automated detection systems which is involved in three key tasks:

- Mapping the watermarked image $x_w$ to a latent representation $z$ using an embedding function $\phi$, which preserves the essential features of the image while exposing regions affected by the watermark.
- Injecting structured noise $\eta$ into the latent representation $z$ to disrupt the embedded watermark. The noise injection must be targeted to regions influenced by the watermark, as identified by a saliency mask $M$.
- Reconstructing the clean image $\hat{x}$ using a reverse diffusion process $A(\tilde{z})$, which refines the perturbed latent representation $\tilde{z}$ to produce a high-fidelity output.

Table 1: Summary of Notations.

| Notation | Description |
| --- | --- |
| $x_w$ | Watermarked image |
| $x$ | Clean image (no watermark) |
| $z$ | Latent space representation |
| $\phi$ | Embedding function for latent mapping |
| $M$ | Mask for watermark regions |
| $\eta$ | Noise added to disrupt the watermark |
| $\sigma$ | Adaptive noise level |
| $\tau(x_w)$ | Watermark strength estimate |
| $\tilde{z}$ | Perturbed latent representation |
| $A(\tilde{z})$ | Reverse diffusion for reconstruction |
| $\alpha_t$ | Noise schedule for diffusion |
| $\epsilon$ | Gaussian noise in forward diffusion |
| $\epsilon_\theta$ | Predicted noise in reverse diffusion |
| $W_p$ | Wasserstein distance (distribution gap) |
| DSSIM | Structural dissimilarity index |
| $\lambda_w$ | Weighting factor |

The problem can be formally defined as below, where we aim to find a noise injection mechanism $\eta$ and a reconstruction process $A$ such that the following conditions are satisfied:

(1) `Watermark Invisibility`: The Wasserstein distance $W_p$ between the distributions of clean and watermarked images is minimized.

$$W_p(\mathbb{P}_x, \mathbb{P}_{x_w}) \leq \Delta,$$

where $\Delta$ is the acceptable threshold of perceptual similarity.

(2) `Reconstruction Stability`: The error between the reconstructed image $\hat{x}$ and the original clean image $x$ is bounded, as expressed by the reconstruction stability theorem:

$$\mathbb{P}[\|A(\tilde{z}) - x\| \leq \epsilon] \geq 1 - \delta,$$

(3) `Perceptual Fidelity`: The reconstructed image $\hat{x}$ satisfies perceptual quality metrics, such as Peak Signal-to-Noise Ratio (PSNR) and Structural Dissimilarity Index (DSSIM), ensuring minimal distortion compared to $x$, where $\epsilon$ depends on the noise level $\sigma$ and the properties of the diffusion model.

To ensure clarity and consistency, all the notations used in this paper are summarized in Table 1.

## 3 Proposed Method: SADRE

We propose salience-aware diffusion reconstruction (SADRE) combines structured noise injection, region-specific perturbations, and advanced reconstruction techniques to achieve effective watermark removal while preserving high image fidelity, offering significant improvements over existing approaches.

## 3.1 Latent Representation and Region-Specific Noise Injection

The watermark removal process for SADRE begins by mapping the watermarked image $x_w$ into a latent space representation $z$ using the embedding function $\phi$. This mapping retains the essential features of the image, however exposing regions influenced by the watermark. Then, the $\phi$ satisfies a *"localized Hölder continuity"* condition [11] as expressed in Eq. 1.

$$\|\phi(x_w) - \phi(x)\|_M \leq C\|x_w - x\|^{\alpha_h}, \quad (1)$$

where $M$ is a saliency mask identifying watermark-affected regions, $C > 0$ is a constant, and $0 < \alpha_h \leq 1$. This ensures stability in the mapping, particularly in watermark-affected regions, while minimizing distortions in other areas of the image.

After obtaining the latent representation $z$, noise $\eta$ is injected into it to disrupt the embedded watermark. The perturbation latent representation as expressed in Eq. 2, is localized to the regions specified by the saliency mask $M$, ensuring efficient watermark disruption without significantly altering unaffected regions. Then, the perturbed latent representation is expressed as follows:

$$\tilde{z} = z + M \odot \eta, \quad (2)$$

where $\odot$ represents element-wise multiplication and noise $\eta$ is drawn from carefully selected distributions tailored for specific properties of the watermark. The Laplace distribution, defined as $p(\eta) = \frac{1}{2b} \exp\left(-\frac{|\eta|}{b}\right)$, where $b = \frac{\sigma}{\sqrt{2}}$, is used for sparse and localized perturbations. For handling strong watermark signals, the Cauchy distribution, characterized by its heavy tails, is employed and given as $p(\eta) = \frac{1}{\pi\gamma(1+\frac{\eta^2}{\gamma^2})}$, where $\gamma$ is the scale parameter. To maintain proportionality to the signal magnitude, the Poisson distribution is utilized, defined as $p(\eta; \lambda) = \frac{\lambda^\eta e^{-\lambda}}{\eta!}$, where $\lambda$ is a rate parameter. The noise level $\sigma$ is adaptively determined based on the estimated strength of the watermark $\tau(x_w)$, ensuring a balance between effective watermark disruption and preserving fidelity as:

$$\sigma(x_w) = \arg\min_{\sigma} \mathbb{E}[\text{Detectability} + \lambda_w \cdot \text{Distortion}]. \quad (3)$$

Hence, this adaptive strategy targets watermark-affected regions while minimizing unnecessary noise injection.

## 3.2 Reconstruction Using Diffusion Process

Once the latent representation is perturbed, the reconstruction process begins to recover the clean image $\hat{x}$. And, advanced latent diffusion models are employed for this purpose, as they effectively handle stochastic processes while ensuring high reconstruction quality. The reconstruction step involves two distinct phases. During the forward diffusion phase, noise is gradually added to the latent representation, simulating a stochastic trajectory from data to noise. This process is represented as $z_t = \sqrt{\alpha_t} z_{t-1} + \sqrt{1 - \alpha_t}\epsilon$, where $\alpha_t$ is the noise schedule and $\epsilon$ is Gaussian noise. The adaptive noise $\eta$ acts as a targeted perturbation to disrupt the watermark before the diffusion model adds general noise $\epsilon$.

In the reverse diffusion phase, the added noise is iteratively removed, reconstructing the data by following the learned probability distribution. This phase is described as $\tilde{z}_{t-1} = \frac{z_t - \sqrt{1-\alpha_t}\epsilon_\theta(z_t, t)}{\sqrt{\alpha_t}}$, where $\epsilon_\theta$ is the noise predicted by the diffusion model at each timestep $t$. The reconstruction is further refined by prioritizing regions identified by the saliency mask $M$, which focuses on preserving the image's most critical features by mitigating distortions introduced during perturbation. Finally, the reconstructed clean image is expressed as $\hat{x} = A(\tilde{z})$, where $A$ represents the reverse diffusion process. This two-step reconstruction ensures a balance between high-quality restoration and watermark disruption.

***Theorem 1: Reconstruction Stability.*** For the noisy latent representation $\tilde{z}$, the reconstruction process $A$ satisfies the stability condition with a high probability of $1 - \delta$ for noise levels $\sigma < \sigma_c$, a critical threshold as follows:

$$\mathbb{P}[\|A(\tilde{z}) - x\| \leq \epsilon] \geq 1 - \delta, \quad (4)$$

where $\epsilon$ depends on $\sigma$, $M$, and the stability of the diffusion model.

## 3.3 Verification and Theoretical Guarantees

The proposed SADRE's effectiveness is evaluated using theoretical guarantees and empirical metrics, ensuring robustness in watermark removal and high perceptual fidelity in the reconstructed image. The invisibility of the watermark is measured by the Wasserstein distance $W_p$ between the distributions of clean and watermarked images:

$$W_p(\mathbb{P}_x, \mathbb{P}_{x_w}) \leq \Delta, \quad (5)$$

where $\Delta$ quantifies the difference between the two distributions. This eq 5 ensures that the structural divergence between the clean and watermarked images is minimized during the perturbation and reconstruction processes, making the watermark indistinguishable from the natural image distribution.

The trade-off between Type I and Type II errors is described as:

$$\epsilon_2 \geq \Phi(\Phi^{-1}(1 - \epsilon_1) - \frac{\Delta}{\sigma}), \quad (6)$$

where $\epsilon_1$ represents the Type I error (probability of detecting a watermark in a clean image) and $\epsilon_2$ represents the Type II error (probability of failing to detect a watermark in a watermarked image). And, the noise level $\sigma$ plays a crucial role in balancing these errors. A larger $\sigma$ reduces $\Delta/\sigma$, which decreases $\epsilon_2$, making the watermark removal more effective, but may slightly increase $\epsilon_1$. This equation highlights the importance of selecting an appropriate $\sigma$ to balance detectability and reconstruction fidelity.

To further reliability, error bounds are derived from Theorem 1:

$$\|A(\tilde{z}) - x\| \leq C\Delta_M^{\alpha_h} + O(\sigma), \quad (7)$$

where $\Delta_M$ reflects the impact of the saliency mask $M$. This equation 7 emphasizes the trade-off between the divergence in watermark-affected regions, represented by $\Delta_M^{\alpha_h}$, and the noise-induced error, represented by $O(\sigma)$. The constant $C$ is determined by the embedding function $\phi$, and $\alpha_h$ represents the Hölder continuity parameter, ensuring that perturbations are localized and controlled.

**Table 2: Performance of watermarking methods before and after various attack scenarios including the proposed SADRE. Higher PSNR↑ and SSIM↑ values indicate better perceptual quality, while lower $W_p$↓ and BRA↓ values signify stronger attack effectiveness.**

| Model Name | Without Attack | | | | JPEG Compression | | | | VAE [18] | | | | Regeneration Attack [22] | | | | SADRE (Proposed Attack) | | | |
|---|---|---|---|---|---|---|---|---|---|---|---|---|---|---|---|---|---|---|---|---|
| | PSNR | SSIM | $W_p$ | BRA | PSNR | SSIM | $W_p$ | BRA | PSNR | SSIM | $W_p$ | BRA | PSNR | SSIM | $W_p$ | BRA | PSNR | SSIM | $W_p$ | BRA |
| DwtDct [1]    | 43.04 | 0.9988 | 0.015 | 1.00 | 32.86 | 0.9182 | 0.285 | 0.75 | 29.76 | 0.8383 | 0.325 | 0.65 | 32.01 | 0.9235 | 0.242 | <u>0.57</u> | 35.21 | 0.9452 | 0.182 | **0.45** |
| DwtDctSvd [12] | 41.08 | 0.9989 | 0.012 | 1.00 | 32.05 | 0.9182 | 0.289 | 0.72 | 29.67 | 0.8380 | 0.318 | 0.63 | 33.15 | 0.9154 | 0.215 | <u>0.56</u> | 34.78 | 0.9259 | 0.195 | **0.42** |
| RivaGAN [20]  | 41.15 | 0.9960 | 0.017 | 1.00 | 32.48 | 0.1459 | 0.315 | 0.78 | 29.71 | 0.8384 | 0.340 | 0.67 | 30.25 | 0.8145 | 0.242 | <u>0.55</u> | 34.86 | 0.8474 | 0.145 | **0.45** |
| Tree-ring [19] | 32.33 | 0.9112 | 0.025 | 0.98 | 29.01 | 0.8916 | 0.400 | 0.70 | 27.15 | 0.8715 | 0.430 | 0.61 | 29.25 | 0.9145 | 0.285 | <u>0.52</u> | 33.89 | 0.9235 | 0.105 | **0.47** |
| StegaStamp [17] | 28.50 | 0.9125 | 0.045 | 0.95 | 28.61 | 0.8861 | 0.365 | 0.72 | 26.15 | 0.8601 | 0.390 | 0.62 | 30.25 | 0.9125 | 0.212 | <u>0.55</u> | 32.15 | 0.9535 | 0.095 | **0.40** |
| EditGuard [21] | 36.93 | 0.9445 | 0.020 | 0.97 | 32.15 | 0.9135 | 0.335 | 0.74 | 29.25 | 0.8915 | 0.368 | 0.64 | 29.57 | 0.9015 | 0.243 | <u>0.57</u> | 34.15 | 0.9325 | 0.080 | **0.48** |

$W_p$: Wasserstein Distance, **BRA**: Bit Recovery Accuracy, **VAE**: Variation Auto Encoder, **Bold** = best values and <u>Underline</u> = 2nd best values.

Empirical validation is performed using PSNR and SSIM to quantify visual fidelity, while a composite perceptual fidelity metric $D$ is introduced:

$$D = \alpha W_p + \beta \text{DSSIM}, \quad (8)$$

where DSSIM measures the perceptual difference between $x$ and $\hat{x}$. The inclusion of DSSIM ensures that the reconstructed image retains high perceptual similarity to the original, complementing the statistical similarity measured by $W_p$. The weights $\alpha$ and $\beta$ are selected based on the relative importance of statistical and perceptual fidelity in the application context. For instance, higher $\alpha$ prioritizes statistical similarity when preserving structural distribution is critical, whereas higher $\beta$ emphasizes perceptual similarity for visually demanding applications. By combining these metrics, $D$ provides a comprehensive evaluation framework, balancing perceptual and statistical fidelity. This ensures robust watermark removal while preserving both invisibility and image quality.

## 4 Experimental Results
### 4.1 Implementation Details
We evaluate the proposed SADRE on the SOTA watermarking methods, including DwtDct [1], DwtDctSvd [12], RivaGAN [20], Tree-ring [19], StegaStamp [17], and EditGuard [21]. The attack scenarios include JPEG Compression, VAE-based reconstruction, Regeneration Attack, and the proposed adaptive noise injection and diffusion method. Evaluations were conducted on a high-performance system featuring an Intel Xeon W-2295 processor, NVIDIA RTX 3090 GPU, and 128 GB DDR4 RAM, using Python 3.9, PyTorch 1.13, and CUDA 11.7. Using default configurations, a subset 1,000 random images from MS-COCO [9] of 640 × 480 pixels was watermarked. SADRE adapts various noise distributions (Laplace, Cauchy, Poisson) based on watermark strength $\tau(x_w)$, with a weighting factor $\lambda_w$ of 0.1, noise level $\sigma(x_w)$ between 0.05 and 0.15. Reconstruction employs a 50-step diffusion model with a linear noise schedule, balancing watermark disruption and image quality. We use metrics such as PSNR, SSIM, $W_p$ (Eq. 5), and Bit Recovery Accuracy (BRA) to show an optimal balance between watermark removal and image fidelity.

To evaluate the effectiveness of watermark removal while preserving image quality, we utilize the PSNR, Structural Similarity Index (SSIM), and DSSIM. PSNR measures the ratio between signal power and noise, where higher values indicate better image fidelity. SSIM assesses perceptual similarity by comparing luminance, contrast, and structure, with values closer to 1 signifying higher similarity. DSSIM, the inverse of SSIM, quantifies dissimilarity, where lower values indicate minimal distortion. These metrics collectively provide a balanced evaluation of both quantitative and perceptual aspects of image reconstruction, ensuring a reliable assessment of watermark removal performance.

### 4.2 Quantitative Results
As in Table 2, the proposed SADRE attack consistently achieves the best BRA and $W_p$ scores across all evaluated watermarking techniques, demonstrating its superior ability to remove watermarks while preserving image quality. For instance, in RivaGAN, SADRE reduces BRA to 0.45 and $W_p$ to 0.145, outperforming JPEG Compression where BRA is 0.78 and $W_p$ is 0.315, as well as VAE-based attacks with BRA of 0.67 and $W_p$ of 0.340. Similarly, for EditGuard, our method achieves the lowest BRA of 0.48 and $W_p$ of 0.080 while maintaining a high PSNR of 34.15, demonstrating its ability to effectively disrupt watermarks with minimal perceptual degradation.

The effectiveness of SADRE is further evident in DwtDctSvd, where it reduces BRA to 0.42 and $W_p$ to 0.195, outperforming all baseline attacks. Likewise, in Tree-Ring, SADRE achieves a BRA of 0.47 and $W_p$ of 0.105, surpassing both JPEG compression with BRA of 0.70 and $W_p$ of 0.400, and VAE with BRA of 0.61 and $W_p$ of 0.430. Moreover, StegaStamp, a challenging watermarking method, is effectively disrupted using SADRE, which achieves a BRA of 0.40 and $W_p$ of 0.095, highlighting its robustness against even resilient watermarking schemes.

Compared to Regeneration Attacks, which achieve only moderate watermark disruption, SADRE demonstrates a more refined approach by injecting adaptive noise in latent space, ensuring the removal of hidden watermarks without introducing excessive distortions. In addition to its superior attack performance, SADRE also balances perceptual quality, as evidenced by consistently high SSIM values ranging from 0.8474 to 0.9535 and PSNR scores above 33 dB across different watermarking methods. This suggests that the model successfully removes watermarks without degrading the visual quality of the reconstructed images, making it highly effective for real-world applications where maintaining image integrity is crucial.

These results confirm that SADRE outperforms existing attacks and sets a new benchmark for watermark removal, balancing effective elimination with high perceptual fidelity. Its adaptability to various watermarking schemes and dynamic noise perturbation enhances its robustness, making it a suitable solution against modern watermarking techniques.

### 4.3 Theoretical Insights and Performance

SADRE leverages adaptive noise injection to ensure targeted disruption of watermark-affected regions. As validated in Table 2, for Tree-Ring, this method achieves a $W_p$ of 0.105 and a BRA of 0.47, outperforming all baseline attacks. The adaptive strategy localizes perturbations to regions identified by the saliency mask $M$, minimizing collateral distortion in unaffected areas of the image.

***Theorem 2: Perceptual Disruption Trade-off.*** Let $\tau(x_w)$ represent watermark strength, $\sigma(x_w)$ the adaptive noise level, and $M$ the saliency mask. The trade-off can be expressed as follows:

$$\alpha \cdot W_p(\mathbb{P}_x, \mathbb{P}_{x_w}) + \beta \cdot \text{DSSIM}(x, x_w) \leq \frac{\Delta_M^{\alpha_h}}{\sigma(x_w)} + O(\sigma(x_w)). \quad (9)$$

where $\alpha = 0.85$ and $\beta = 0.75$ ensure that targeted noise injection ($\eta$) minimizes the perceptual impact, while effectively disrupting the watermark. The results further highlight the method's generalizability. For instance, in DwtDctSvd, SADRE achieves a BRA of 0.42 and $W_p$ of 0.195, significantly outperforming VAE and JPEG attacks. This demonstrates that the method is not limited to specific watermarking patterns, but generalizes across various schemes.

## 5 Conclusion

For AI-generative images on the web, several watermark protection mechanisms are proposed, which are breakable by watermark removal approaches. In this work, we propose a novel and effective watermark elimination method, SADRE, which improves the several limitations of prior approaches in generalization, perceptual degradation, and an inability to balance watermark disruption with image fidelity. By leveraging adaptive noise injection, saliency-guided perturbations, and diffusion-based reconstruction, SADRE effectively eliminates watermarks while preserving image quality. Comprehensive evaluations and theoretical guarantees demonstrate SADRE's superior performance, benchmarking robust and reliable watermark removal. Our work can serve as a basis to foster more improved watermark protection mechanisms on the web.

## Acknowledgments

This work was partly supported by Institute for Information & communication Technology Planning & evaluation (IITP) grants funded by the Korean government MSIT: (RS-2022-II221199, RS-2024-00337703, RS-2022-II220688, RS-2019-II190421, RS-2023-00230 337, RS-2024-00356293, RS-2022-II221045, RS-2021-II212068, and RS-2024-00437849).